\documentclass[10pt,twocolumn,letterpaper]{article}

\usepackage{iccv}
\usepackage{times}
\usepackage{epsfig}
\usepackage{graphicx}
\usepackage{amsmath}
\usepackage{amssymb}

\usepackage[breaklinks=true,bookmarks=false]{hyperref}

\iccvfinalcopy 

\def\iccvPaperID{****} 

\ificcvfinal\fi

\begin{document}

\title{Deepfake Detection Scheme Based on Vision Transformer and Distillation}

\author{Young-Jin Heo, Young-Ju Choi, Byung-Gyu Kim\\
Sookmyung Women's University\\
IVPL\\
{\tt\small \{yj.heo, yj.choi, bg.kim\}@ivpl.sookmyung.ac.kr}

\and
Young-Woon Lee\\
Sunmoon University\\
IVPL\\
{\tt\small yw.lee@ivpl.sookmyung.ac.kr}
}

\maketitle
\ificcvfinal\thispagestyle{empty}\fi
\begin{abstract}
   Deepfake is the manipulated video made with a generative deep learning technique such as Generative Adversarial Networks (GANs) or Auto Encoder that anyone can utilize. Recently, with the increase of Deepfake videos, some classifiers consisting of the convolutional neural network that can distinguish fake videos as well as deepfake datasets have been actively created. However, the previous studies based on the CNN structure have the problem of not only overfitting, but also considerable misjudging fake video as real ones. In this paper, we propose a Vision Transformer model with distillation methodology for detecting fake videos. We design that a CNN features and patch-based positioning model learns to interact with all positions to  find the artifact region for solving false negative problem. Through comparative analysis on Deepfake Detection (DFDC) Dataset, we verify that the proposed scheme with patch embedding as input outperforms the state-of-the-art using the combined CNN features. Without ensemble technique, our model obtains 0.978 of AUC and 91.9 of f1 score, while previous SOTA model yields 0.972 of AUC and 90.6 of f1 score on the same condition.
\end{abstract}

\section{Introduction}
Deepfake is a combination of `Deep learning' and `fake', which refers to the technique of changing a source person on a target video. This technology makes the source person seem to be doing the actions or saying things of the target person by superimposing the source person as a target person as a deep learning model.

However, cases of abuse such as fake news and revenge porn have emerged as a social issue. Because it is a technology that can confuse people, some technologies and datasets for detecting fake videos have been studied in response. Fake videos can be detected by locating and detecting artificial parts within frames or between frames. Networks that search for factitious parts in a frame are mainly composed of the CNN by considering spatial characteristics. In the DFDC full dataset \cite{dolhansky2020deepfake} released by Facebook AI, EfficientNetwork-b7 \cite{Selim} became the SOTA model. In this paper, we find fake videos using these spatial characteristics.

The Transformer model has currently been actively researched in the field of NLP and Computer VIsion \cite{vaswani2017attention}. In BERT \cite{devlin2018bert}, only the encoder part of Trasnformer was used to produce state-of-the-art models on GLUE, MultiNLI, SQuAD v1.1 and SQuAD v2.0. 
Another model, GPT \cite{brown2020language}, \cite{radford2019language}, \cite{radford2018improving}, uses only the transformer's decoder and boasts high performance in the field of document generation.

The first vision transformer \cite{dosovitskiy2020image} by the Google, originally has been used in NLP task. It contributed in classification task. Another Transformer network called DeiT \cite{touvron2020training} by facebook used a distillation methodology and gave higher accuracy than VIT (Vision Transformer) even fewer datasets. From this observation, we add a distill token to the VIT and consider patch embedding and CNN features together for the model input to create a more generalized model for deepfake detection. Based on this, we can expect that this model is more robust on deepfake detection. This paper is organised as follows: In Section 2, we describe the related works. The proposed scheme is explained in Section 3. Section 4 will give experimental results and analysis. Finally, we will make a concluding remark in Section 5.

\section{Related Work}
\subsection{Face synthesis}
Image sysnthesis network are divided into two approaches: Generative Adversarial Netowrk (GAN) \cite{goodfellow2014generative} and Variational AutoEncoder (VAE) \cite{kingma2013auto}. The GAN networks have more application models than VAE. Especially cGAN \cite{mirza2014conditional}, WGAN \cite{arjovsky2017wasserstein}, WGAN-GP \cite{wei2018improving}, PGGAN \cite{karras2017progressive}, DCGAN \cite{radford2015unsupervised}, DiscoGAN \cite{kim2017learning}, CycleGAN \cite{zhu2017unpaired}, StarGAN \cite{choi2018stargan}, and StyleGAN \cite{karras2019style} are well-known. Currently, StyleGAN-V2 \cite{karras2020analyzing} is a leader in facial synthesis as well as many synthetic programs being shared as open source anyone can use. The first network to produce deepfake videos is the VAE. As suggested by deepfake autoencoder (DFAE), source images which are the faces to be synthesized to the target are trained by each two VAE models and then switched the last decoder part that identifies who it is. As a result, images are changed from the face of the target video to the face of the source video. 

According to \cite{nguyen2019deep}, this approach applies to deepfake program such as DeepFaceLab \cite{deepfaceLab}, DFaker \cite{DFaker}, and DeepFaketf \cite{DeepFaketf}.
Face Swapping GAN (FSGAN) is subject agnostic face swapping and reenactment without requiring training on those faces \cite{nirkin2019fsgan}. They used a face mixing network to blend the two faces seamlessly while preserving the target skin color and lighting conditions. 

\subsection{Deepfake detection}
Most configured models in deepfake detection is based on the CNN structure. There are two approaches for discriminating deepfake videos. One is to exploit unnatural spatial properties within one frame of video as an image unit, and the other is to exploit temporal properties to find unnaturalness between video frames. 

Montserrat detects the space-time awkwardness by putting the frames of the video into the efficientnet and each feature into the Gated Recurrent Unit (GRU) \cite{montserrat2020deepfakes}. Similarly, G{\"u}era use CNN to extract frame-level features and train a RNN that learns to classify fake videos \cite{guera2018deepfake}. In addition, Unlike previous studies using CNN and RNN networks to find out spatio-temporal properties, de Lima \cite{de2020deepfake} use 3DCNN to detect them at once. They use I3D \cite{carreira2017quo}, R3D \cite{hara2017learning}, MC3 \cite{tran2018closer} because of higher performing network architectures. As the method of optical flow based cnn, Amerini proposed a optical flow field to exploit possible inter-frame dissimilarities \cite{amerini2019deepfake}.

To detect the spatial of manipulated in the face, Li \cite{li2018exposing} use CNN model such as VGG16 \cite{simonyan2014very}, ResNet50, ResNet101, and ResNet152 \cite{he2016deep}. Nguyen proposed a capsule network that can detect various kinds of deepfakes \cite{nguyen2019use}. they use the features pretrained by VGG16 and suggested Capsule-Forensics architecture.
Classical classification method using the SVM was proposed by Yang \cite{yang2019exposing} and Guarnera used K-nearest neighbors and linear discriminant analysis \cite{guarnera2020deepfake}.
\section{Proposed Deepfake Detection Algorithm}
\begin{figure*}[!t]
\centering
\includegraphics[width=5in]{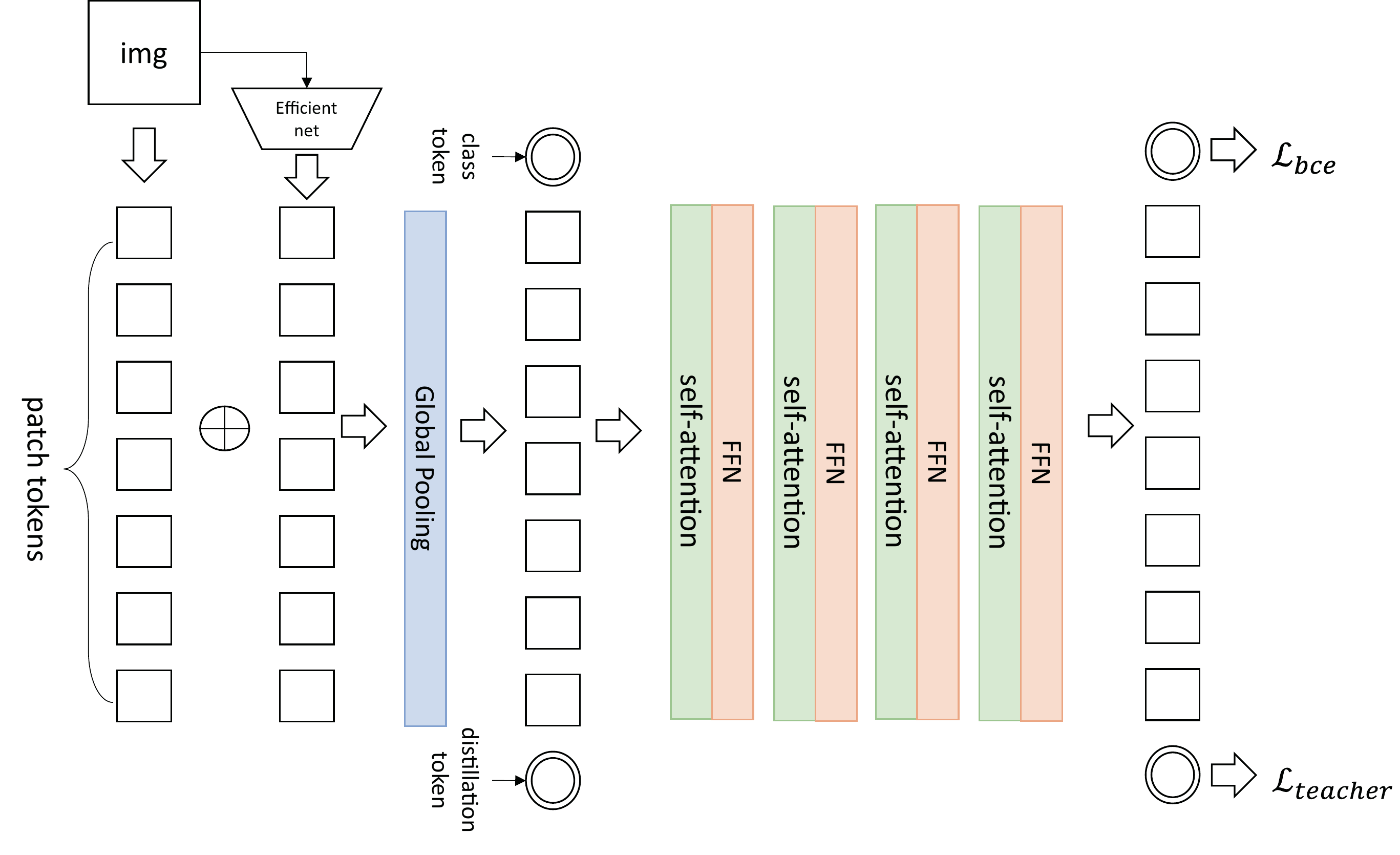}
\caption{The proposed overall deepfake detection network. The image split into patches and pass EfficientNet \cite{tan2019efficientnet}. We got (Batch, N, embedding features), (Batch, M, embedding features) respectively. These tokens are concatenated, through global pooling, and fed to transformer encoder. Encoder consist of Multi-Self Attention (MSA) and 2 layers of GeLU function. We add distillation token which is trained by the teacher network.}
\label{fig:1}
\end{figure*}
In this section, we describe the Vision Transformer architecture for deepfake detection. Our baseline follows Vision Transformer network with adding distillation token. Input sequences are combined patch embedding with the CNN features. The entire network is shown in Figure \ref{fig:1}. We introduce the role and characteristic of Vision Transformer for deepfake detection in Section 3.1, Specifically, we illustrate that how the input is consisted in Section 3.2 and distillation method with teacher network in Section 3.3.

\subsection{Base network architecture}
We overview the Vision Transformer \cite{dosovitskiy2020image} and recognize its effectiveness in the field of deepfake detection. Transformer was originally used for NLP task but recently, there are many attempts to apply on image modeling \cite{girdhar2019video}, \cite{neimark2021video}, \cite{liu2021swin}. The Vision Transformer only has encoder similar to BERT, which has position information and embedding sequences. 

In \cite{dosovitskiy2020image}, before the multi head self attention layers, image $x \in \mathbb{R}^{(H\times W\times C)}$ split into patch $x_p \in \mathbb{R}^{(H/P \times W/P \times E)}$ by learnable embedding, where (H,W) is resolution of the image, $C$ is channel, $P$ is patch size and $E$ is the number of embedding feature. All patches are flattened by linear projection and add to the position embedding equal in $(H/P \times W/P)$. The Transformer encoder consists of multiheaded self attention (MSA) and multi layer perceptron (MLP). MLP contains two layers with a Gaussian Error Linear Unit (GELU) non-linearity \cite{dosovitskiy2020image}. We follow the MSA, MLP function as encoder to discriminate robustly fake video detection.

Sequences of feature vectors include all part of the image. In addition, an encoder refer to all of the sequences of split patches. The previous CNN structure employed attention only activated part of face and could not refer to other distant position. However, input sequences depend on global information and this point can reduce the overfitting problem in transformer. Also, we find out interesting result that transformer makes relatively fair classification of real and fake videos, rather than skewed to either side, unlike the previous CNN model.

\subsection{Combination of patch embedding and CNN features}
We introduce the input tokens before feeding to encoder, unlike the original oneself from input vectors of vision transformer. We define $\mathbf{Z}_p = (x^1_p\boldsymbol{E},x^2_p\boldsymbol{E}, \cdots, x^N_p\boldsymbol{E})$ and $\mathbf{Z}_f = f(x) = (x^1_f, x^2_f, \cdots, x^M_f)$, where $x_p$ is a patch, $\boldsymbol{E}$ is a learnable embedding, N is exponential the number of split patches, M is the number of CNN features and $f(\cdot)$ is a CNN model. Thus, $\mathbf{Z}_f$ means feature of CNN model and we use $f$ as EfficientNet. 

These features are combined as $\mathbf{Z}_p \oplus \mathbf{Z}_f$ and applied global pooling. Min Lin suggested that the global average pooling was more interpretable between feature maps and categories \cite{lin2013network}. We represent $\mathbf{Z}_{p\oplus f} = globalpooling(\mathbf{Z}_p \oplus \mathbf{Z}_f)$ as input vectors, by $N+M$ to $N$(vectors number). As a result, we consider not only the main part features of face, but also all parts correlation. By using thie approach, we can get better AUC and f1 score than only using patch embedding or CNN features.

\subsection{Distillation method and teacher network}
We add class token and distillation token to input $\mathbf{Z}_p \oplus \mathbf{Z}_f$, so we define final input $\mathbf{Z}_0 = [x_{class};\mathbf{Z}_P\oplus \mathbf{Z}_f;x_{distillation}] + \boldsymbol{E}_{pos}$, where $x_{class}$ and $x_{distillation}$ is token for training by label and teacher network, $\boldsymbol{E}_{pos}$ is the learnable position embedding.
Finally, we can define a the set train loss as:
\begin{flalign}
    \mathcal{L}_{fake} &= \lambda \mathcal{L}_{BCE}(Z_{c_{fake}}, y) + && \nonumber \\ 
    &(1-\lambda)  \mathcal{L}_{BCE}(Z_{d_{fake}}, \sigma(Z_{t_{fake}})), &&
\end{flalign}
\begin{flalign}
    \mathcal{L}_{real} &= \lambda \mathcal{L}_{BCE}(Z_{c_{real}}, y) + &&\nonumber \\ 
    &(1-\lambda)  \mathcal{L}_{BCE}(Z_{d_{real}}, \sigma(Z_{t_{real}})), &&
\end{flalign}
\begin{flalign}
    &\mathcal{L}_{train}= (\mathcal{L}_{fake} + \mathcal{L}_{real})/2, &
\end{flalign}
where ($Z_{t_{fake}}$, $Z_{t_{real}}$) are the logits of the teacher model for fake prediction and real prediction,  ($Z_{d_{fake}}$, $Z_{d_{real}}$) and ($Z_{c_{fake}}$, $Z_{c_{real}}$) are the logits of the distillation tokens and the class tokens for fake prediction and real prediction. We set $\lambda$ by $\frac{1}{2}$, binary cross entropy ($\mathcal{L}_{BCE}$) on the labels $y$ and $\sigma$ the sigma function.

In \cite{touvron2020training} by Facebook AI, a distillation method has the effect of preventing overfitting by expanding the range of weights of labels. Also, when teacher network was the CNN model, transformer produced the best result than other models. 

Therefore, we choose the teacher network as EfficientNet which is the state of the art model on the DFDC dataset in deepfake detection. Each class token and distillation token represents the probability that the video is fake. The distillation tokens are used instead of class token when testing. It outperforms class token on test dataset.

\section{Experiments}
In this section, we describe the dataset and detail of parameter. Also, we compare to the SOTA model, representing condition of performance measurements. We will explain why we use the DFDC dataset in Section 4.1. Also, we describe the parameter setting and configuration environments required in the training process in Section 4.2, and analyze the experimental results in Section 4.3 compared to the SOTA model.
\subsection{DFDC Dataset}
In kaggle contest\footnote {https://www.kaggle.com/c/deepfake-detection-challenge}, they opened DFDC Preview dataset \cite{dolhansky2019deepfake} and later, facebook AI opened full DFDC dataset \cite{dolhansky2020deepfake}. The DFDC dataset is the largest and public-available deepfake dataset, including about 100,000 total videos produced by generative adversarial network (GAN).

In \cite{dolhansky2020deepfake}, the face swap datasets are divided into three generations. First-generation datasets such as DF-TIMIT \cite{korshunov2018deepfakes}, UADFV \cite{yang2019exposing}, and FaceForensics++DF (FF++DF) \cite{rossler2019faceforensics++} have $10^4 ~ 10^6$ frames and up to 5000 videos. Second-generation datasets are Celeb-DF \cite{li2020celeb} and DFDC Preview \cite{dolhansky2019deepfake}. DFDC full dataset is third-generation and has 128,154 total videos and 104,500 unique fake videos.

For these reasons, we choose the largest deepfake dataset and compare the performance with the state-of-the-art (SOTA) on the DFDC full dataset. In the analysis of Dolhansky \cite{dolhansky2020deepfake}, the submitted best model has 0.734 AUC in the private test set. Also, the higher average precision of submitted models (such as \cite{Selim}, \cite{Hanqing}, \cite{Azat}, \cite{Jing}, \cite{James}) in DFDC dataset, the better performance in real video \cite{dolhansky2020deepfake}. Therefore, if the performance is good in the DFDC dataset, the results can be generalized in real video.

\begin{figure}[!t]
  \centering
  \includegraphics[width=7cm, height=4.0cm]{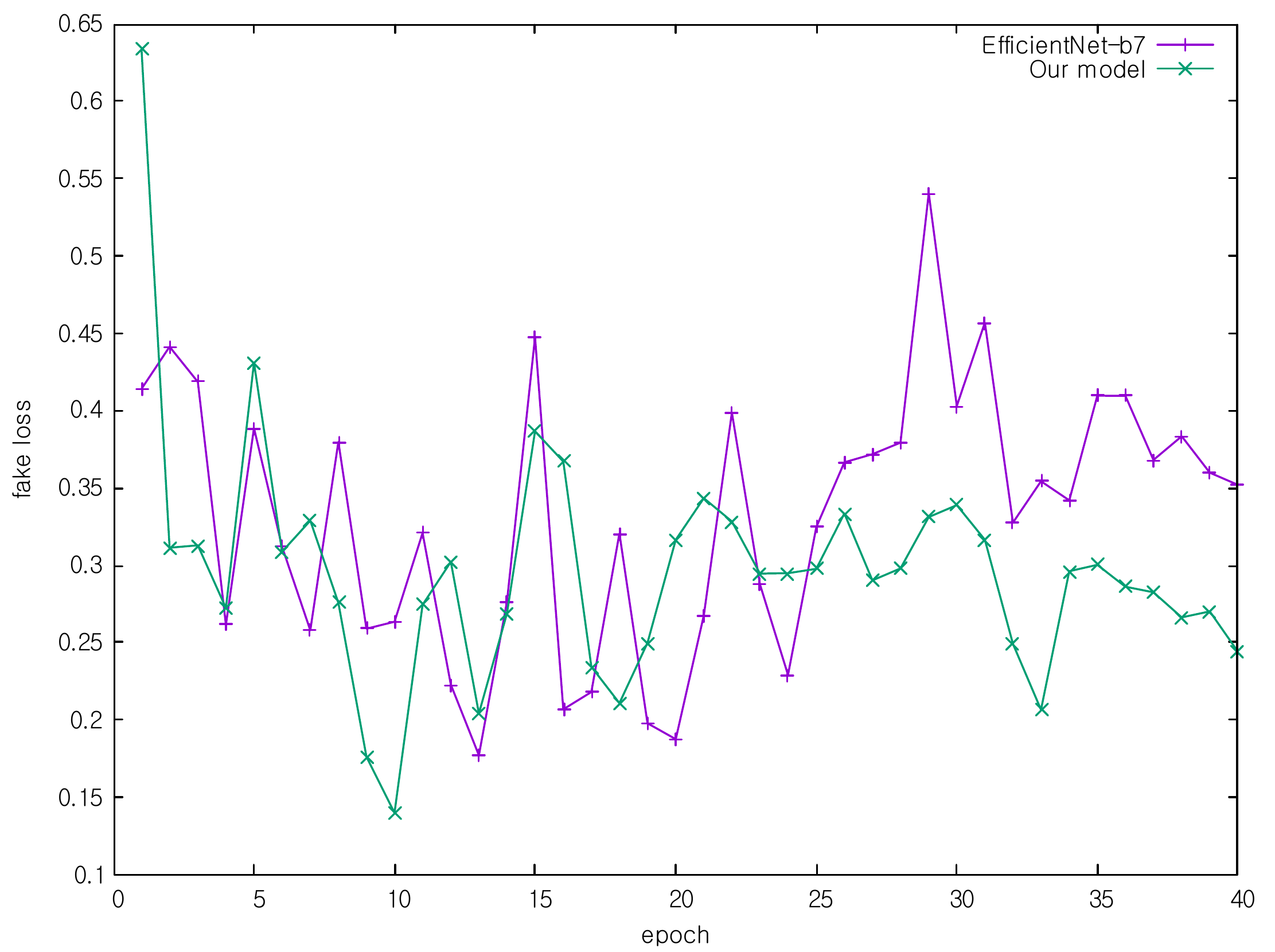}
  \centerline{(a)}
  \includegraphics[width=7cm, height=4.0cm]{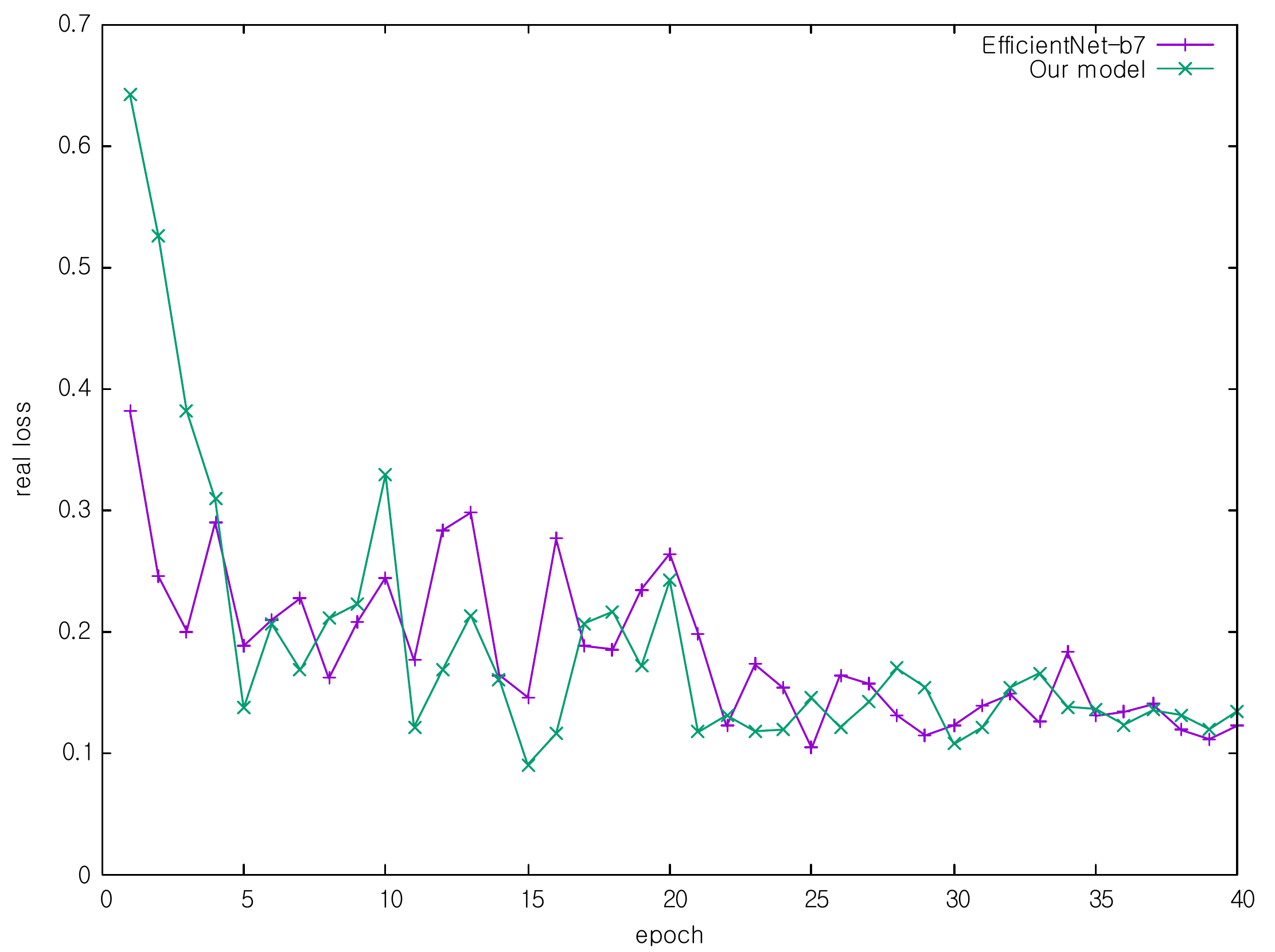}
  \centerline{(b)}
  \includegraphics[width=7cm, height=4.0cm]{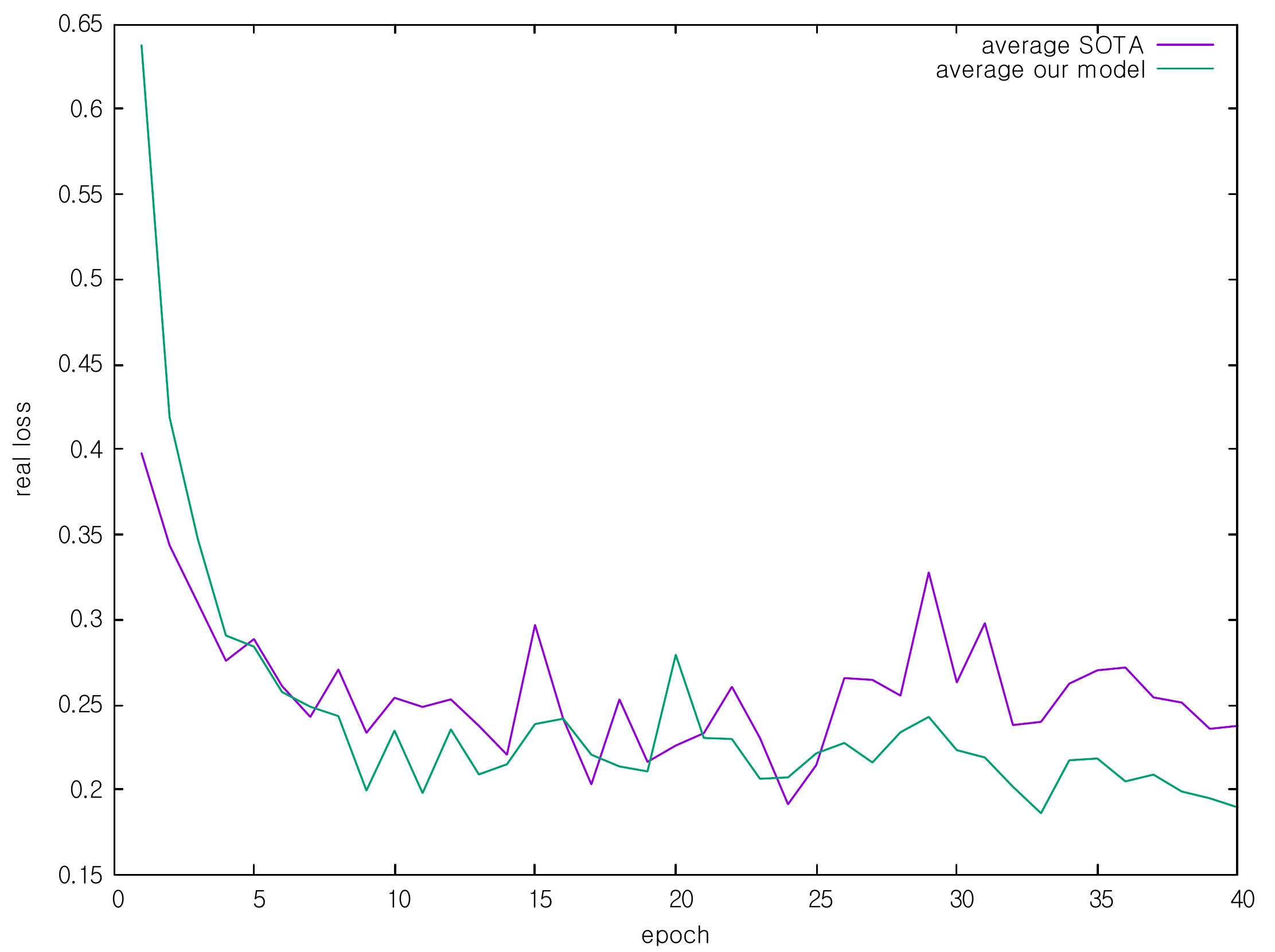}
  \centerline{(c)}
  \caption{The results of loss in validation DFDC dataset: (a) represents the loss for fake video, (b) represents the loss for real video, and (c) represents the average loss. The green plot means our model's loss and the purple plot means the SOTA model's loss. There is little difference in loss of real video, but much difference in fake video.}
  \label{fig:2}
\end{figure}

\begin{figure}[!t]
  \centering
  \includegraphics[width=7cm, height=4cm]{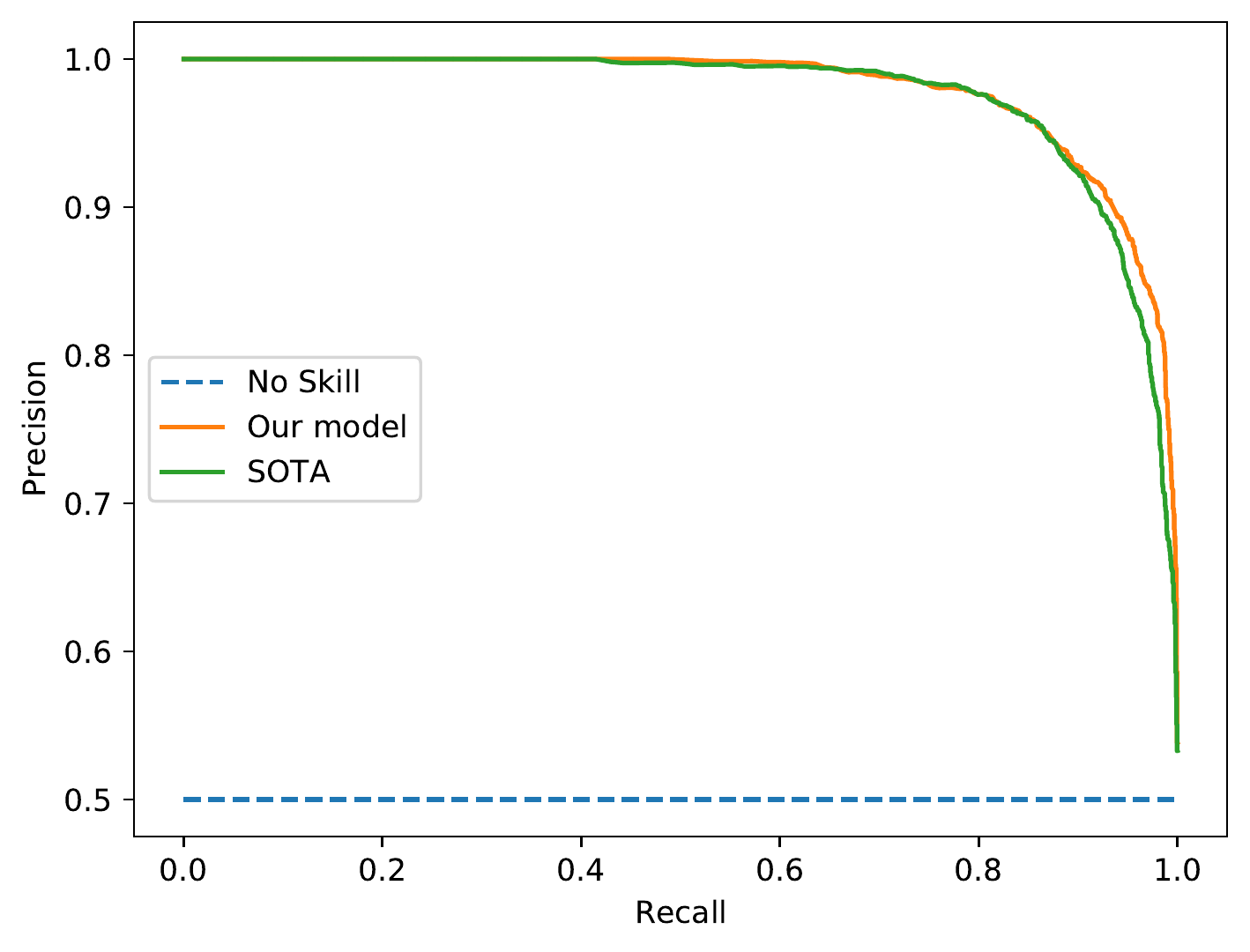}
  \includegraphics[width=7cm, height=4cm]{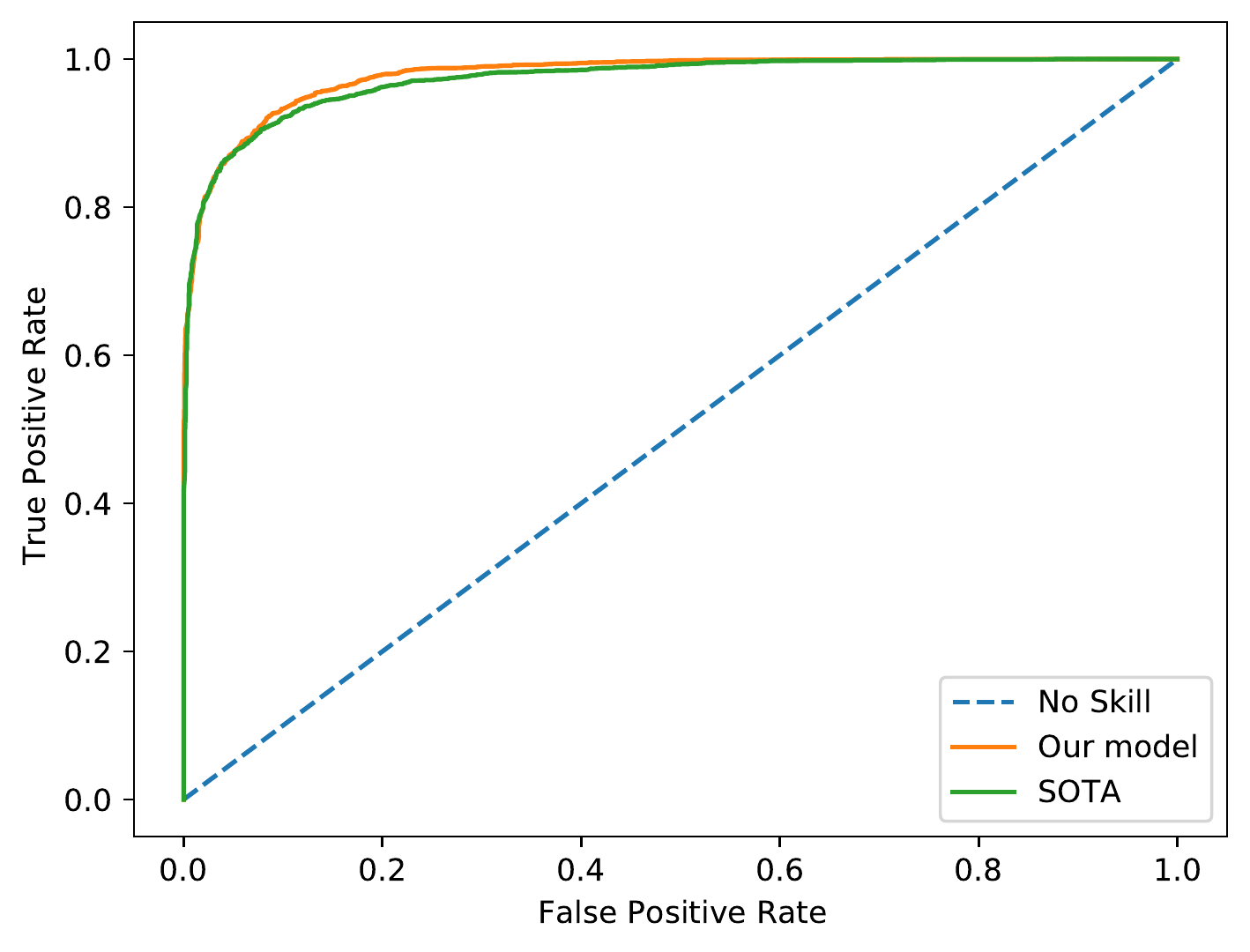}
  \caption{ROC-AUC curve. Orange is the curve of our model and green is the curve of SOTA model. Our model has larger area than the compared SOTA.}
  \label{fig:3}
\end{figure}

\subsection{Training Detail}
{\bf Pre-processing} : We used a face detector as the Multi task Cascaded Convolutional Networks (MTCNN) and crop all the frames to 384x384. We augmented our training data using Albumentations \cite{buslaev2020albumentations}. In addition, we cutout and dropout the part of the image simliar to \cite{Selim}. Our pre-processing procedure was based on Selim\cite{Selim}.

{\bf Training} : Our patch size for the embedding features is 32 and the embedding dimension is 1024. We initialize our transformer and Efficientnet-b7 model by the pre-trained model. We set the transformer 16-heads and 24-layers which is identical to the large VIT default model. Also, our teacher network used a pre-trained one by Efficientnet-B7 \cite{Selim} on the DFDC dataset. We used a distillation token when testing.   

{\bf Parameter} : training and test were done on a V100-GPU machine with a batch size of 12 for training. We used an SGD optimizer with an initial learning rate of 0.01 and a different learning rate reduction policy which is a step-based method. The training epoch is 40, batches per epoch are 2500 and it takes 2 days on single V100-GPU. 

For classification, we propagate logit values backward from the transformer model using the binary cross entropy. We tested on the public available DFDC 5000 test dataset. The f1 score was measured for comparison with the SOTA model with the 0.55 of threshold value.


\subsection{Performance analysis}
We compare our model to the state-of-the-art model \cite{Selim}. We trained our model by train dataset and chose the model weights with the lowest loss in the validation set. In Figure \ref{fig:2}, we compare the validation loss with  the SOTA model \cite{Selim} on real and fake video respectively. This graph shows that our model is a more robust classifier on fake videos. Also, Real loss is similar, but the overall average loss was lower. The test loss is defined as:
\begin{eqnarray}
\boldsymbol{LogLoss} = -\frac{1}{n} \sum^n_{i=1}[y_ilog(\hat{y}_i + (1-y_i)log(1-\hat{y}_i)], 
\end{eqnarray}
where $n$ is the number of videos being predicted, $\hat{y}_i$ is the predicted probability of the video being FAKE, $y_i$ is 1 if the video is FAKE, 0 if REAL. We get $\hat{y}_i$ by distillation tokens.

In addition, our model's ROC-AUC curve (0.978) has a larger area than the SOTA model (0.972) in Figure \ref{fig:3}. It represents the proposed classifier is a more robust than the SOTA model on fake videos, because the precision is higher than the SOTA model as the recall is close to 1. 

\begin{figure}[!t]
  \centering
  \includegraphics[scale=0.35]{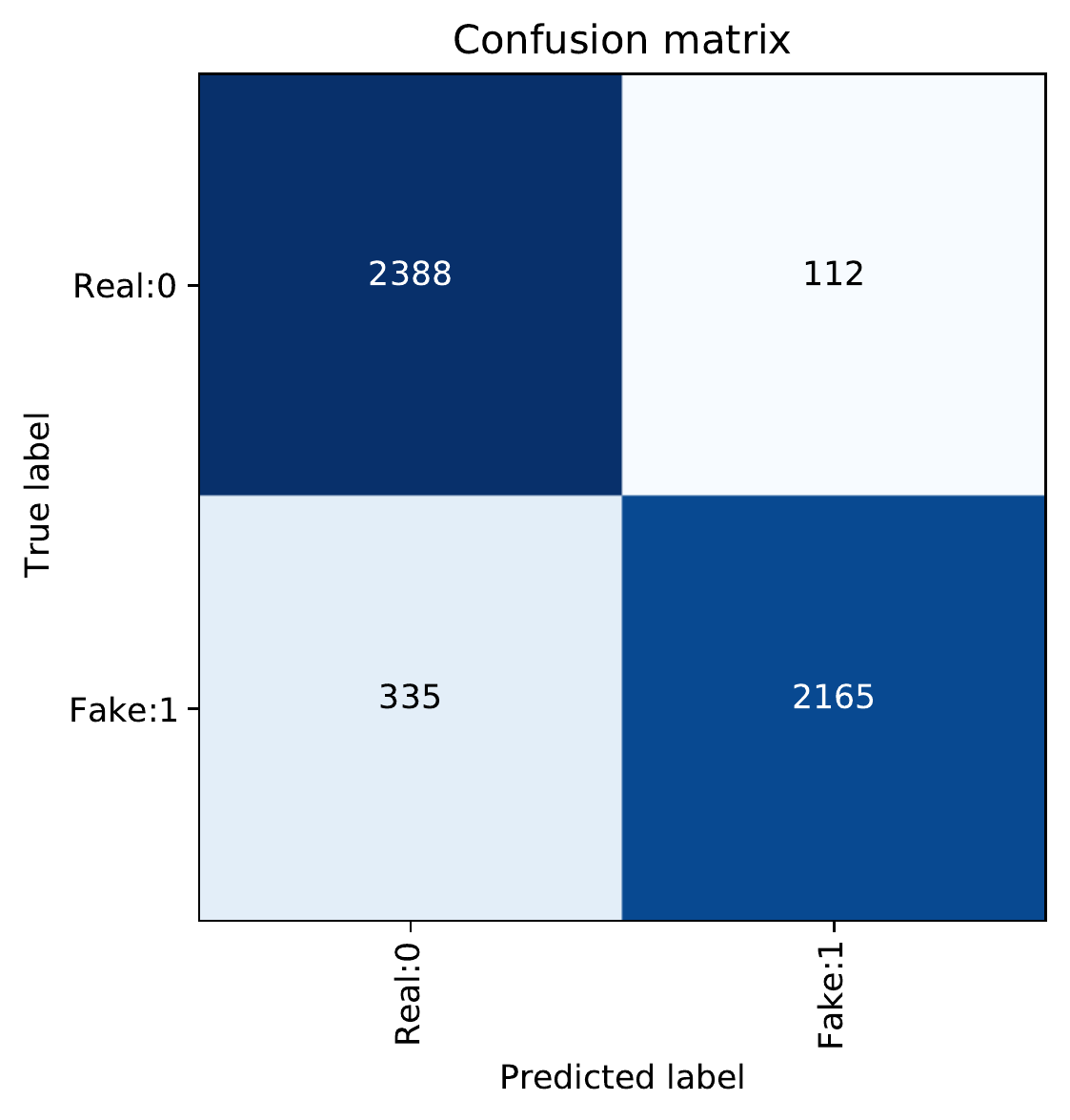}
  \includegraphics[scale=0.35]{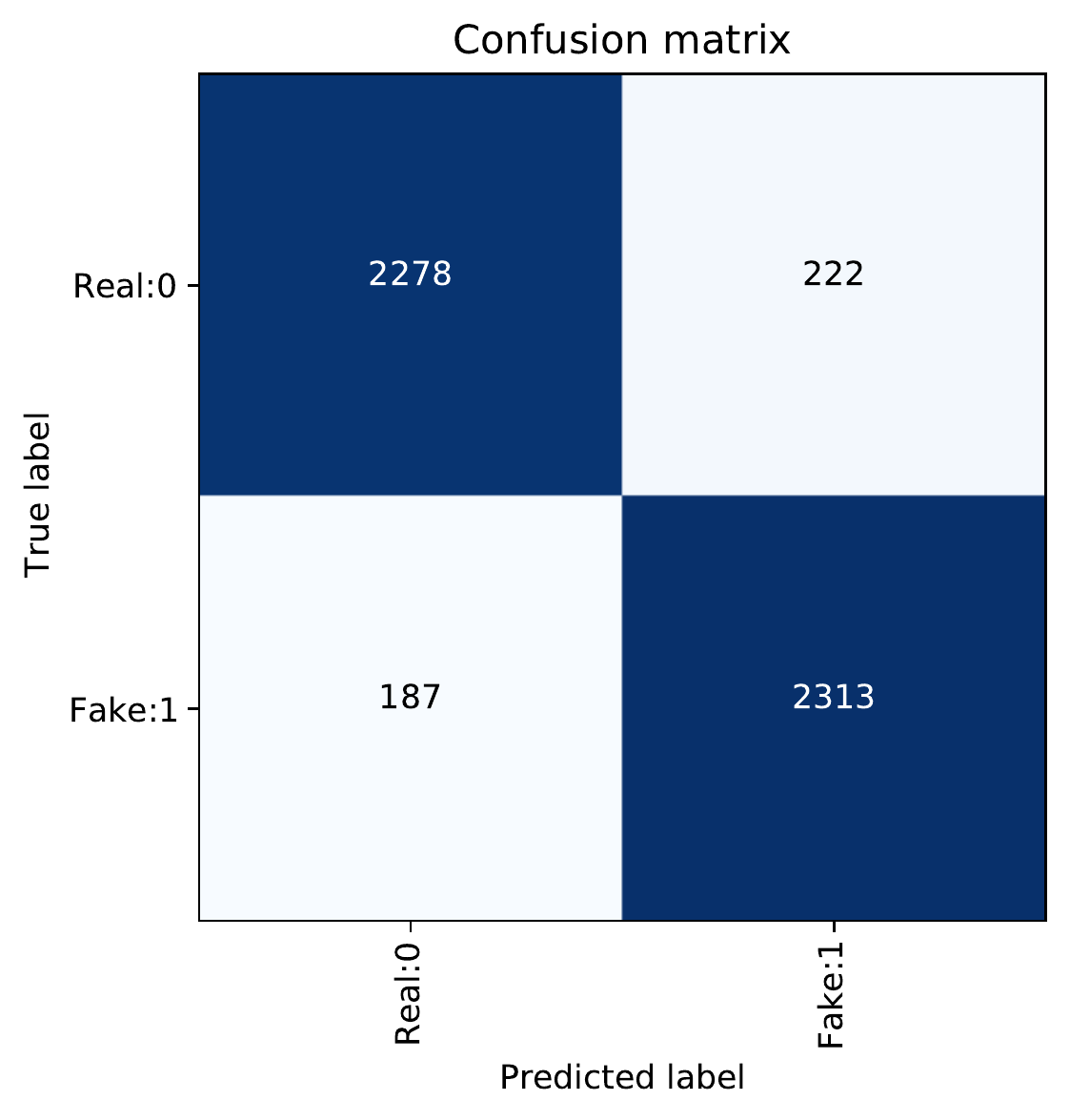}
  \caption{We set the threshold 0.55 which is probability of fake videos. Top-right, top-left, bottom-right, bottom-left means false-positive, true-negative, true-positive, false-negative in order. Confusion Matrix of the left side is previous SOTA model's prediction and the right side is our results. We can see that our model predict clearly on fake videos.}
  \label{fig:4}
\end{figure}

To verify robust classification, the confusion matrix was obtained by setting the threshold of 0.55, which represents the probability of fake video. It represents the  predicted number of video according to each label in Figure \ref{fig:4}. Each model's result of false negative was 335 and 187. Thus, we can see that the proposed model is robust in fake videos and the f1 score was 91.9, which was higher than 90.6 of the SOTA model.

\begin{figure}[!t]
  \centering
  \includegraphics[scale=0.5]{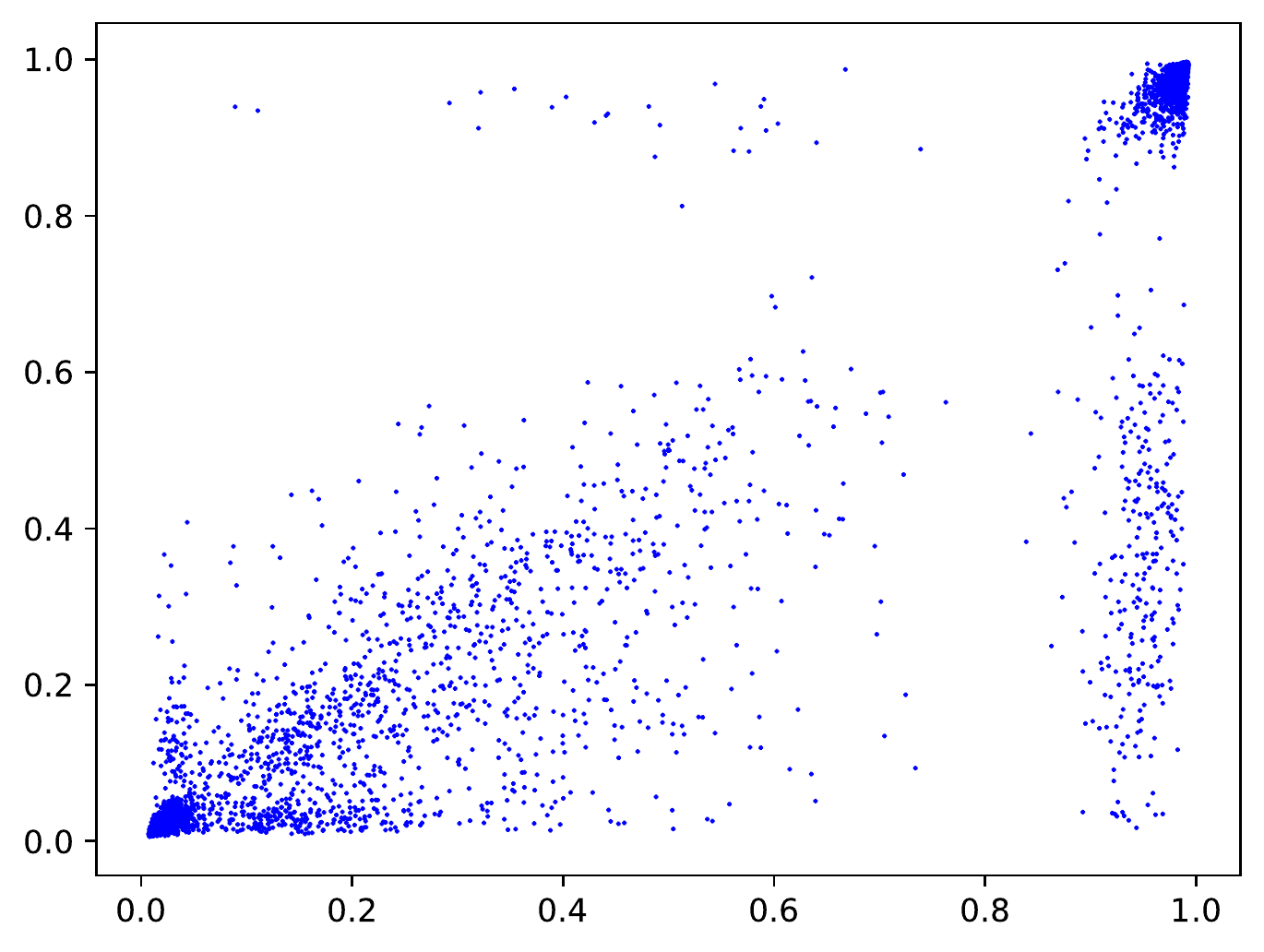}
  \caption{Correlation of prediction values between our model and SOTA. The x-axis is the predicted probability of our model whether the video is fake or not, and the y-axis is the predicted value of the SOTA model. We note that our model classifies fake videos clearly but the SOTA model is not confident about many fake videos.}
  \label{fig:5}
\end{figure}

To figure out the correlation of predictions between the SOTA model and our model, we scatter all the fake video prediction probabilities for 5000 videos. In Figure \ref{fig:5}, when the x-axis is the value predicted by our model and the y-axis is the value predicted by SOTA, the correlated data are concentrated on (0,0) and (1,1). We noted that, in the part that our model predicts as 1 (predicted as fake video), the probability values of the SOTA model are evenly spread. This condition describes that our model can predict a fake video more clearly.


\section{Conclusion}
In this paper, we have proposed a robust Vision Transformer model for Deepfake detection. The proposed scheme was combined vision transformer and EfficientNet in the patch embedding level and distillation technique. We demonstrated the efficiency of the robust Vision Transformer model compared to the previous SOTA model, EfficientNet, which consisted of a CNN network. We observed 0.972 of AUC for the SOTA model and 0.978 for our model under conditions of the same environment without ensemble. For f1 score, the proposed scheme gave the better performance as 91.9 while the STOA model achieved 90.6 in same threshold condition threshold of (0.55).
{\small
}

\end{document}